\documentclass[runningheads,a4paper]{llncs}

\usepackage{amssymb}
\usepackage{amsmath}
\usepackage{graphicx}
\usepackage{multirow}
\usepackage{subfigure}
\usepackage{authblk}

\usepackage{url}
\urldef{\mailstud}\path|afmollah@gmail.com|
\urldef{\mailguides}\path|{subhadip,mnasipuri}@cse.jdvu.ac.in |

\newcommand{\keywords}[1]{\par\addvspace\baselineskip
\noindent\keywordname\enspace\ignorespaces#1}

\begin{document}

\mainmatter  

\title{Computationally Efficient Implementation of Convolution-based Locally Adaptive Binarization Techniques}

\titlerunning{Efficient Implementation of Adaptive Binarization Techniques}

%
%

%

\author{Ayatullah Faruk Mollah\inst{1} \and Subhadip Basu\inst{2} \and Mita Nasipuri\inst{2}}

\institute{
    School of Mobile Computing and Communication, \\Jadavpur University, India \\
    \email{\mailstud}
    \and
    Department of Computer Science \& Engineering,\\Jadavpur University, India\\
    \email{\mailguides}
}

%

\toctitle{Lecture Notes in Computer Science}
\tocauthor{Authors' Instructions}
\maketitle

\begin{abstract}
One of the most important steps of document image processing is binarization. The computational requirements of locally adaptive binarization techniques make them unsuitable for devices with limited computing facilities. In this paper, we have presented a computationally efficient implementation of convolution based locally adaptive binarization techniques keeping the performance comparable to the original implementation. The computational complexity has been reduced from $O(W^2 N^2)$ to $O(WN^2)$ where $W\times W$ is the window size and $N\times N$ is the image size. Experiments over benchmark datasets show that the computation time has been reduced by 5 to 15 times depending on the window size while memory consumption remains the same with respect to the state-of-the-art algorithmic implementation.
\end{abstract}

\section{Introduction}

Document image binarization is an extensively studied topic over the past decades. It is one of the most important steps of any document processing systems. It can be defined as a process of converting a multi-chromatic digital image into a bi-chromatic one. A multi-chromatic image also called as color image consists of color pixels each of which is represented by a combination of three basic color components viz. red ($r$), green ($g$) and blue ($b$). The range of values for all these color components is 0-255. So, the corresponding gray scale value $f(x,y)$ for a pixel located at $(x,y)$ may be obtained by using Eq. \ref{eqGrayConv}.

\begin{equation}\label{eqGrayConv}
    f(x,y)=w_r\times r(x,y)+w_g\times g(x,y)+w_b\times b(x,y)
\end{equation}

\noindent
where $w_r$= 0.299, $w_g$= 0.587 and $w_b$= 0.114. As $\sum w_i$ =1, the range of $f(x,y)$ is also 0-255. So, a gray scale image can be represented as a matrix of gray level intensities $F_{M\times N}$ = $[f(x,y)]_{M\times N}$ where $M$ and $N$ denote the number of rows i.e. the height of the image and the number of the columns i.e. the width of the image respectively. Similarly, a binarized image $G_{M\times N}$ can be represented as $[g(x,y)]_{M\times N}$ such that $g(x,y)\in$ \{0, 255\}.

Techniques developed so far for document image binarization are categorized into two types - global binarization techniques and locally adaptive binarization techniques. In the first case, pixels constituting the image are binarized with a single threshold $T$ as shown in Eq. \ref{EqBnz}. A number of such techniques \cite{Abutaleb_Auto}-\cite{Otsu_AThr} have been developed, of which Otsu's technique \cite{Otsu_AThr} has been found to be the best in a study conducted by Trier et al. \cite{Trier_Eval}-\cite{Trier_Goal}.

\begin{equation}\label{EqBnz}
    g(x,y)=
    \begin{cases}
        0 & \text{if $f(x,y)< T$}\\
        255 & \text {Otherwise}
    \end{cases}
\end{equation}

Global binarization techniques in general produce good results for noise free and homogeneous document images of good quality. But, it fails to properly binarize the images with uneven illumination and noise. Locally adaptive binarization techniques evolved to overcome this problem by binarizing pixels with pixel specific threshold $t(x,y)$ as shown in Eq. \ref{EqAdapBnz}.

\begin{equation}\label{EqAdapBnz}
    g(x,y)=
    \begin{cases}
        0 & \text{if $f(x,y)< t(x,y)$}\\
        255 & \text {Otherwise}
    \end{cases}
\end{equation}

Quite a good number of such adaptive techniques have been found in the literature \cite{Bernsen_Dyn}-\cite{Trier_Impr}. Among these techniques, the best one has been found to be Niblack's one \cite{Niblack_An} in the same study by Trier et al. \cite{Trier_Eval}-\cite{Trier_Goal}. Later, more advanced techniques have been designed and some of them have been reported in \cite{Sauvola_Adap}-\cite{Shin_Block}. Sauvola's \cite{Sauvola_Adap} text binarization method (TBM) is one of them. This method calculates $t(x,y)$ from mean $m(x,y)$ and standard deviation $s(x,y)$ of the gray levels of the pixels within a window around the subject pixel $(x,y)$ as described in Eq. \ref{EqSauvBnz}.

\begin{equation}\label{EqSauvBnz}
    t(x,y)=m(x,y)[1+k(\frac{s(x,y)}{R}-1)]
\end{equation}

\noindent
where $k$ is a positive constant and $R$ is the dynamic range of standard deviation. A good number of the locally adaptive binarization techniques including Sauvola \cite{Sauvola_Adap} are convolution based. As a result, the computational complexity and computation time of such techniques are very high. So far, binarization techniques are evaluated on the basis of binarization accuracy only \cite{Gatos_DIBCO09}-\cite{Pratikakis_HDIBCO10}. But, study on computational requirements of algorithms is also required especially for real time systems and low-resourceful computing devices such as cell-phones, Personal Digital Assistants (PDA), iPhones, iPod-Touch, etc.

The present work is an attempt to reduce the computational complexity of convolution based binarization techniques while retaining comparable accuracies. The computational complexity of the global binarization technique is usually $O(N^2)$ where $N\times N$ is the image size. In case of convolution based binarization techniques, selection of threshold for each pixel requires computation of mean and standard deviation of the gray level intensities of the surrounding pixels within the window. So, computation of individual threshold value for each pixel has a complexity of $O(W^2)$ where $W\times W$ is the window size and overall complexity is $O(W^2 N^2)$ for an image of $N\times N$ pixels.

As the study \cite{Dunlop_Challenges} shows, handheld/mobile devices may not be capable of running algorithms of $O(W^2 N^2)$ within affordable time. Even, the time taken for such algorithms on desktop computers may lead to dissatisfaction to many. In \cite{Shafait_Eff}, Shafait et al. have suggested an implementation for such algorithm with computational complexity of $O(N^2)$. They proposed a faster calculation of $m(x,y)$ and $s(x,y)$ using integral images, but at the cost of 5 to 6 times more memory. In the present work, we have proposed a novel implementation of the convolution based binarization algorithms, which has a computational complexity of $O(WN^2)$ and does not require any additional memory. Experimental results on publicly available standard datasets and our own dataset have also been presented.

\section{Present Work}
The computation of mean \emph{m(x,y)} and standard deviation \emph{s(x,y)} for each pixel \emph {(x,y)} is the most time consuming operation in convolution based locally adaptive binarization techniques. If a window of size ${W \times W}$ pixels is taken around a pixel \emph {(x,y)}, then the set of window pixels, S, will have  ${W \times W}$ number of elements. The performance of such methods is heavily dependent on the size of the window. Window size is decided on the basis of pattern stroke and pattern size. It cannot be made arbitrarily small.

A possible way to reduce the execution time of such binarization methods is to reduce the number of pixels in S by considering only the pixels which effectively contribute in computation of mean and variance within the window. In our present work, we have tried to reduce this number by sampling pixels from S following some geometrical order to form a reduced set S$'\subset$ S.

\begin{figure}
  \centering
  \includegraphics[width=300pt]{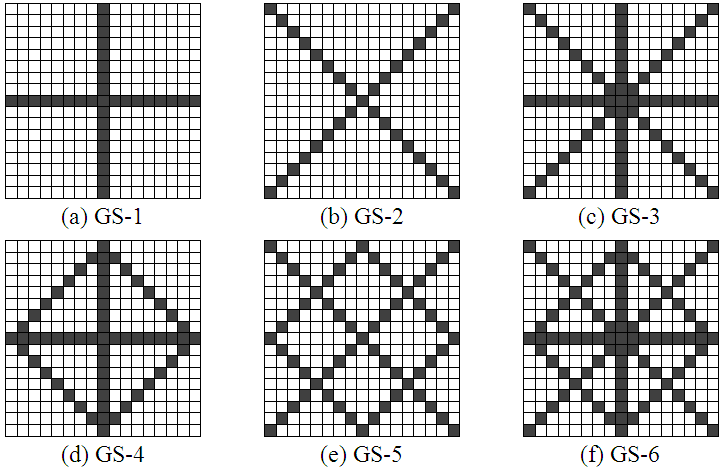}\\
  \caption{Various geometric structures for selection of representative pixels in a window}
  \label{fig_GS}
\end{figure}

Different geometric structures can be defined to select the contributive pixels. A few of such geometric structures have been shown in Fig. \ref{fig_GS}. S$'$ contains pixels corresponding to black boxes marked in the geometric structures of Fig. \ref{fig_GS}. It may be observed that in S$'$, the number of foreground pixels is much lesser than that of the background pixels for the windows of same size around both foreground and background pixels. The mean and standard deviation computed from S$'$ are denoted as $m'(x,y)$ and $s'(x,y)$ respectively.


It is evident that S$'$ is a very small subset of S for all possible geometric structures of Fig. \ref{fig_GS}. Also, in this context, the formulation of Sauvola's method is as given in Eq. \ref{EqnSauvBnzMod}.

\begin{equation}
\label{EqnSauvBnzMod}
    t'(x,y)= m'(x,y)[1+k(\frac{s'(x,y)}{R'}-1)]
\end{equation}

\noindent
where $t'(x,y)$ denotes the threshold calculated from the reduced set S$'$ for the pixel $(x,y)$, $R'$ is the dynamic range of $s'(x,y)$ and $k$ is a positive constant.

\section{Experimental Results}

The proposed implementation has been tested on the printed as well as handwritten images used for benchmarking the performance of various binarization techniques in the recent Document Image Binarization Contest (DIBCO) 2009 \cite{url_dibco09} and Handwritten Document Image Binarization Contest (H-DIBCO) 2010 \cite{url_hdibco10}. It has also been tested on our own dataset (CMATERdb-6) as well. This dataset contains 5 representative images. The first one is of a handwritten Bengali document in which the texts on the rear side are visible from the front side and the image is unevenly illuminated. The second image is of an old historical printed Bengali document. The third one is an image of printed English text and has been captured from a notice board by a cell-phone camera. The fourth image is of an old printed document having texts of multiple fonts and font-sizes. The last one is a cell-phone camera captured business card image.

Ground truth data for the DIBCO and H-DIBCO datasets are publicly available. We have prepared the ground truth data for the images of CMATERdb-6 dataset. The original and ground truth images of this dataset are publicly available for research purposes at \cite{url_cmaterdb}. These three datasets together have 25 representative images containing various kinds of degradations and deformations. The results obtained with the proposed implementation have been compared with the original/algorithmic implementation of Sauvola's binarization method, since it is one of the best convolution-based binarization methods. Results obtained with Niblack as well as Otsu's binarization technique have also been given.

\subsection{Performance Analysis}
Current implementation incepts from a conjecture that the threshold value \emph{t$'$(x,y)} obtained from \emph{m$'$(x,y)} and \emph{s$'$(x,y)} are not considerably different from \emph{t(x,y)} computed from \emph{m(x,y)} and \emph{s(x,y)}. As a result, the performance remains comparable. The binarized result obtained with the new threshold $t'(x,y)$ may not be exactly same with that obtained with \emph{t(x,y)}, but experiments show that the results obtained with the presented technique serves the purpose of binarization very well.

Comparing the output images obtained using various geometric structure of Fig.\ref{fig_GS}(a-f) with their ground truth images, we find the number of true positives (\emph{TP}), number of true negatives (\emph{TN}), number of false positives (\emph{FP}) and the number of false negatives (\emph{FN}). The definition for F-Measure (\emph{FM}) in terms of Recall rate (\emph{R}) and precision rate (\emph{P}) has been given in Eq. \ref{eqn_FM}.

\begin{equation}\label{eqn_FM}
    FM=\frac{2\times R \times P}{R +P}
\end{equation}

\noindent
where $R=\frac{TP}{TP+FN}$ and $P=\frac{TP}{TP+FP}$. In an ideal situation i.e. when the output image is identical with the ground truth image, $R$, $P$ and $FM$ should be all 100\%. While calculating the F-Measure, the best combination of window size ($W$) and $k$ has been considered in all cases.

Table \ref{tab_FM4DIBCO} shows F-Measures achieved with Otsu, Niblack, Sauvola and proposed implementations of Sauvola's method for DIBCO image dataset. It contains 5 (1-5) printed and 5 (6-10) handwritten images. Bold cells represent the highest F-Measure achieved for the corresponding image. It may be noted that the highest mean F-Measure (91.13\%) has been achieved with GS-3. Moreover, mean F-Measures achieved with GS-4, GS-5 and GS-6 are greater than that of Sauvola's method. F-Measure with GS-2 is equal to that of Sauvola.

\begin{table}
  \centering
  \caption{F-Measures achieved with different techniques/implementations for DIBCO images (\textbf{Bold} cells represent the highest F-Measure for the corresponding image)}
  \label{tab_FM4DIBCO}
  \setlength{\tabcolsep}{5pt}
  \begin{tabular} {|c|c|c|c|c|c|c|c|c|c|}
    \hline
    \multicolumn{1}{|c|}{\multirow{2}{*} {Image}} &
    \multicolumn{9}{|c|}{F-Measures (\%)}      \\ \cline{2-10}
    \multicolumn{1}{|c|}{}                        &
    Otsu & Niblack & Sauvola & GS-1 & GS-2 & GS-3 & GS-4 & GS-5 & GS-6   \\ \cline{1-10}

    1 &	91.06 &	88.12 &	91.64 &	91.18 &	91.83 &	91.88 &	91.61 &	\textbf{91.91} &	91.71 \\ \hline
    2 &	\textbf{96.56} &	94.76 &	96.39 &	95.58 &	96.27 &	96.16 &	96.14 &	96.40 &	96.25 \\ \hline
    3 &	\textbf{96.71} &	88.65 &	95.82 &	94.90 &	95.83 &	95.58 &	95.68 &	95.96 &	95.77 \\ \hline
    4 &	82.59 &	90.29 &	92.93 &	91.70 &	93.02 &	92.88 &	92.56 &	\textbf{92.93} &	92.86 \\ \hline
    5 &	89.58 &	85.59 &	89.81 &	88.49 &	89.85 &	89.36 &	89.33 &	\textbf{89.87} &	89.62 \\ \hline
    6 &	90.85 &	\textbf{92.70} &	92.34 &	91.64 &	91.96 &	92.15 &	91.77 &	91.78 &	91.98 \\ \hline
    7 &	86.15 &	75.02 &	86.65 &	\textbf{89.64} &	88.99 &	89.41 &	89.21 &	89.06 &	89.23 \\ \hline
    8 &	84.11 &	88.19 &	87.99 &	88.08 &	88.02 &	\textbf{89.08} &	88.74 &	88.46 &	88.92 \\ \hline
    9 &	40.56 &	86.20 &	88.62 &	88.09 &	88.65 &	\textbf{89.24} &	88.52 &	88.79 &	88.99 \\ \hline
   10 &	28.04 &	85.62 &	85.38 &	\textbf{86.36} &	83.13 &	85.59 &	85.13 &	85.05 &	84.81 \\ \hline
Mean &	78.62 &	87.51 &	90.76 &	90.57 &	90.76 &	\textbf{91.13} &	90.89 &	91.02 &	91.01 \\ \hline
  \end{tabular}
\end{table}

Similar to Table \ref{tab_FM4DIBCO}, Table \ref{tab_FM4HDIBCO} shows F-Measures achieved with H-DIBCO images. It contains 10 representative handwritten document images. Proposed implementations have achieved highest F-Measures for 6 images out of 10. The implementation referred to as GS-3 alone has yielded 3 highest F-Measures. Although, the mean F-Measure is highest in case of Sauvola, F-Measures of the proposed implementations are close to that.

\begin{table}
  \centering
  \setlength{\tabcolsep}{5pt}
  \caption{F-Measures achieved with different techniques/implementations for H-DIBCO images (\textbf{Bold} cells represent the highest F-Measure for the corresponding image)}
  \label{tab_FM4HDIBCO}
  \begin{tabular} {|c|c|c|c|c|c|c|c|c|c|}
    \hline
    \multicolumn{1}{|c|}{\multirow{2}{*} {Image}} &
    \multicolumn{9}{|c|}{F-Measures (\%)}      \\ \cline{2-10}
    \multicolumn{1}{|c|}{}                        &
    Otsu & Niblack & Sauvola & GS-1 & GS-2 & GS-3 & GS-4 & GS-5 & GS-6   \\ \cline{1-10}

   1  &  \textbf{91.47} &	90.98 &	91.23 &	89.39 &	90.82 &	89.71 &	90.28 &	91.37 &	90.57 \\ \hline
   2  &  88.18 &	88.46 &	89.03 &	\textbf{89.86} &	88.32 &	89.47 &	89.53 &	88.64 &	89.26 \\ \hline
   3  &  84.36 &    81.78 &	\textbf{85.64} &	84.16 &	85.01 &	84.36 &	84.45 &	85.15 &	84.63 \\ \hline
   4  &  85.62 &	89.80 &	89.67 &	89.29 &	89.82 &	\textbf{89.84} &	89.45 &	89.60 &	89.69 \\ \hline
   5  &  88.28 &	84.57 &	92.26 &	92.91 &	93.20 &	\textbf{93.51} &	93.19 &	93.23 &	93.38 \\ \hline
   6  &  80.38 &	84.38 &	84.09 &	84.04 &	83.77 &	84.11 &	\textbf{84.42} &	84.33 &	84.33 \\ \hline
   7  &  90.12 &	89.57 &	90.87 &	90.76 &	90.69 &	\textbf{91.13} &	90.84 &	90.79 &	91.09 \\ \hline
   8  &  85.68 &	88.32 &	88.23 &	87.27 &	88.01 &	88.29 &	88.30 &	88.12 &	\textbf{88.32} \\ \hline
   9  &  81.28 &	\textbf{88.43} &	88.42 &	88.40 &	87.88 &	87.88 &	87.85 &	87.92 &	87.92 \\ \hline
  10  &  79.25 &	87.67 &	87.60 &	\textbf{87.90} &	86.00 &	85.91 &	85.54 &	85.67 &	85.66 \\ \hline
Mean  &  85.46 &	87.40 &	\textbf{88.70} &	88.40 &	88.35 &	88.42 &	88.39 &	88.48 &	88.49 \\ \hline

  \end{tabular}
\end{table}

\begin{table}
  \centering
  \setlength{\tabcolsep}{5pt}
  \caption{F-Measures achieved with different techniques/implementations for CMATERdb6 images (\textbf{Bold} cells represent the highest F-Measure for the corresponding image)}
  \label{tab_FM4CMATERDB6}
  \begin{tabular} {|c|c|c|c|c|c|c|c|c|c|}
    \hline
    \multicolumn{1}{|c|}{\multirow{2}{*} {Image}} &
    \multicolumn{9}{|c|}{F-Measures (\%)}      \\ \cline{2-10}
    \multicolumn{1}{|c|}{}                        &
    Otsu & Niblack & Sauvola & GS-1 & GS-2 & GS-3 & GS-4 & GS-5 & GS-6   \\ \cline{1-10}

    1 & 88.00 &	89.71 &	89.98 &	90.10 &	90.09 & \textbf{90.16} &	90.05 &	89.96 &	90.14 \\ \hline
    2 &	88.88 &	88.68 &	89.04 &	88.97 &	\textbf{89.07} &	89.05 &	89.06 &	89.07 &	89.07 \\ \hline
    3 & 91.93 &	92.89 &	93.41 &	93.00 &	93.37 &	\textbf{93.46} &	93.29 &	93.38 &	93.46 \\ \hline
    4 &	\textbf{99.04} &	95.64 &	98.02 &	97.01 &	97.42 &	97.90 & 97.80 &	97.82 &	97.94 \\ \hline
    5 &	91.06 &	90.94 &	91.65 &	91.40 &	91.63 &	91.70 &	\textbf{91.81} &	91.77 &	91.80 \\ \hline
    Mean &	91.78 &	91.57 &	92.42 &	92.10 &	92.32 &	92.45 &	92.40 &	92.40 &	\textbf{92.48} \\ \hline
  \end{tabular}
\end{table}
Table \ref{tab_FM4CMATERDB6} shows the F-Measures achieved with CMATERdb-6 images. It is noteworthy that highest mean F-Measure i.e. 92.48\% has been achieved for GS-6 whereas the mean F-Measure for Sauvola's method is 92.42\%. It may also be noted that Otsu's global binarization method has given the highest F-Measure for the fourth image.

\begin{figure}
  \includegraphics{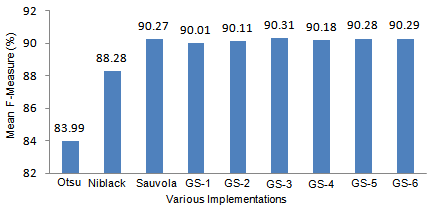}\\
  \caption{Mean F-Measures computed for all images of the 3 benchmarking datasets with various techniques and implementations}
  \label{fig_mean_fm4allsets}
\end{figure}

\begin{figure}
\centering
\mbox{
    \subfigure[]{\setlength\fboxsep{0pt} \setlength\fboxrule{0.5pt} \fbox{\includegraphics[height=1.5in]{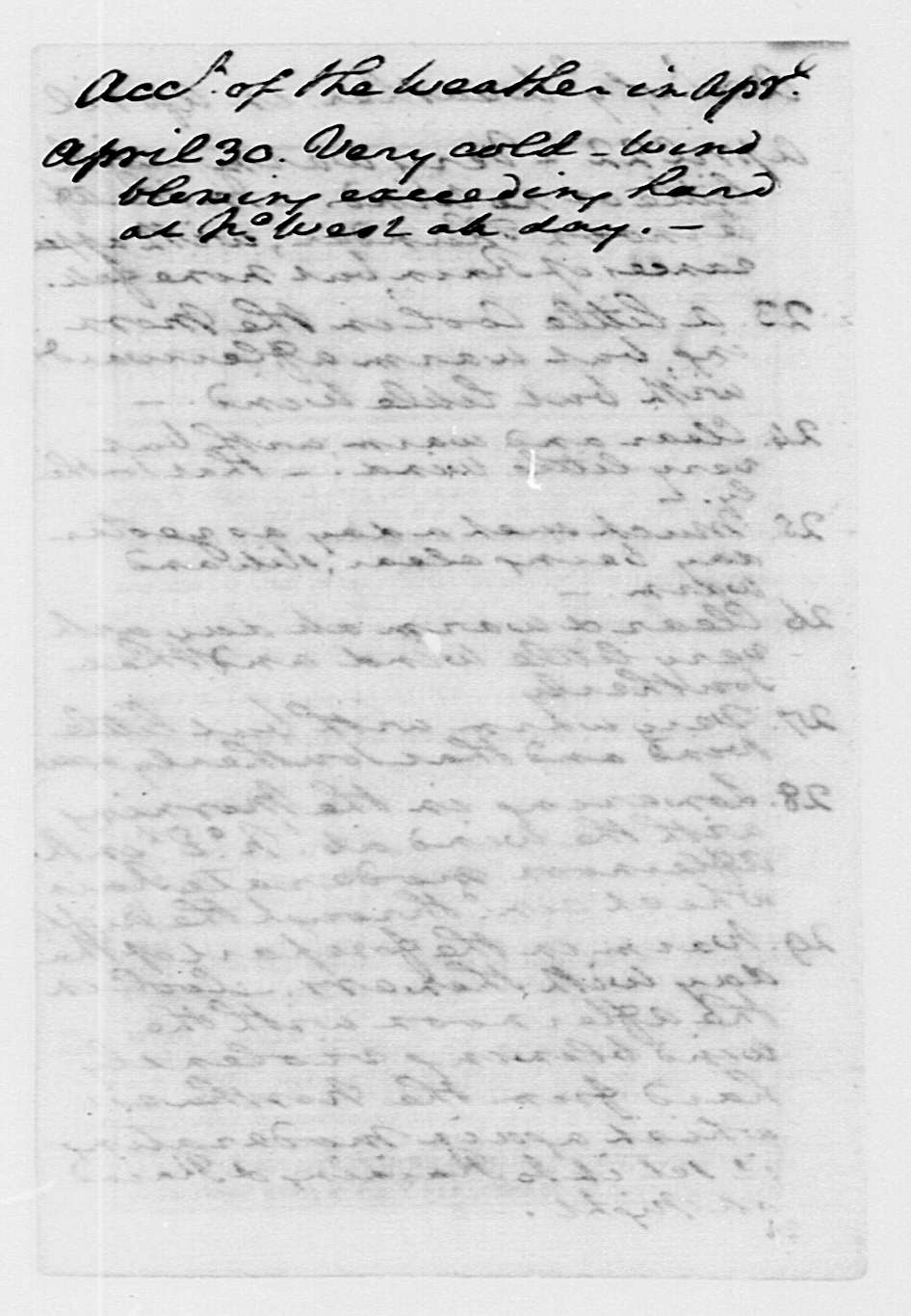}}}

    \subfigure[]{\setlength\fboxsep{0pt} \setlength\fboxrule{0.5pt} \fbox{\includegraphics[height=1.5in]{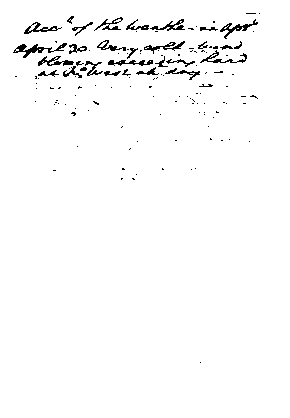}}}

    \subfigure[]{\setlength\fboxsep{0pt} \setlength\fboxrule{0.5pt} \fbox{\includegraphics[height=1.5in]{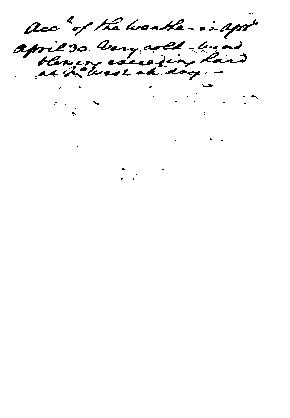}}}

    \subfigure[]{\setlength\fboxsep{0pt} \setlength\fboxrule{0.5pt} \fbox{\includegraphics[height=1.5in]{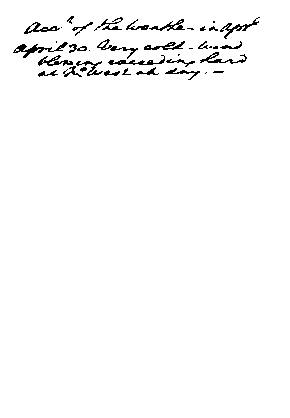}}}
}
\mbox{
    \subfigure[]{\setlength\fboxsep{0pt} \setlength\fboxrule{0.5pt} \fbox{\includegraphics[height=0.55in]{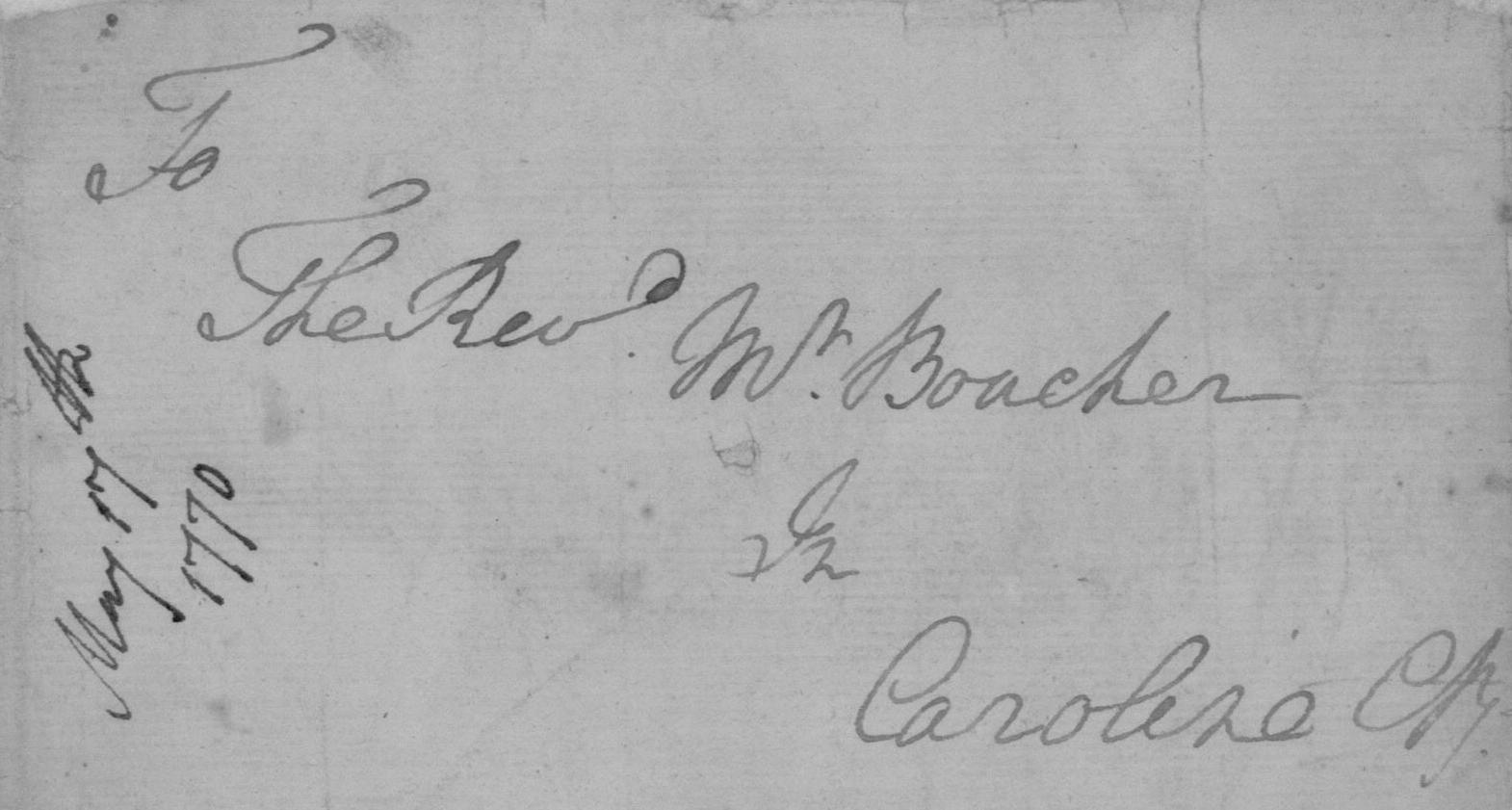}}}

    \subfigure[]{\setlength\fboxsep{0pt} \setlength\fboxrule{0.5pt} \fbox{\includegraphics[height=0.55in]{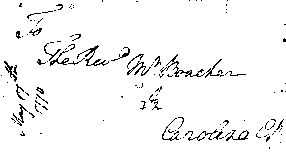}}}

    \subfigure[]{\setlength\fboxsep{0pt} \setlength\fboxrule{0.5pt} \fbox{\includegraphics[height=0.55in]{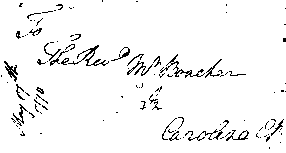}}}

    \subfigure[]{\setlength\fboxsep{0pt} \setlength\fboxrule{0.5pt} \fbox{\includegraphics[height=0.55in]{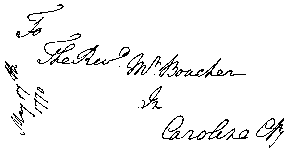}}}
}
\mbox{
    \subfigure[]{\setlength\fboxsep{0pt} \setlength\fboxrule{0.5pt} \fbox{\includegraphics[height=0.5in]{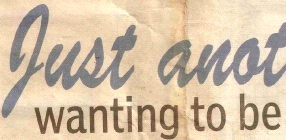}}}

    \subfigure[]{\setlength\fboxsep{0pt} \setlength\fboxrule{0.5pt} \fbox{\includegraphics[height=0.5in]{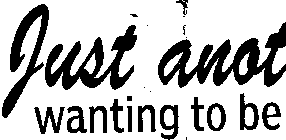}}}

    \subfigure[]{\setlength\fboxsep{0pt} \setlength\fboxrule{0.5pt} \fbox{\includegraphics[height=0.5in]{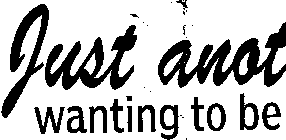}}}

    \subfigure[]{\setlength\fboxsep{0pt} \setlength\fboxrule{0.5pt} \fbox{\includegraphics[height=0.5in]{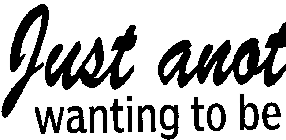}}}
}

\caption{Sample images and binarized results with various techniques. (a,e,i) Image \#7 of \cite{url_dibco09}, \#2 of \cite{url_hdibco10} and \#4 of \cite{url_cmaterdb} respectively, (b,f,j) Binarized images with Sauvola's method, (c,g,k) Binarized images with GS-3, GS-1 and GS-3 respectively, (d) Ground truth images}
\label{fig_benchmark_bnz}
\end{figure}

 A comparison of the mean F-Measures achieved for all 25 images of the 3 datasets with various techniques has been shown in Fig. \ref{fig_mean_fm4allsets}. The highest F-Measure i.e. 90.31\% is achieved with GS-3. F-Measures achieved with all present implementations are greater than that of Niblack. Three implementations viz. GS--1, GS--2 and GS--4 have yielded F-Measures slightly less than that of Sauvola and the remaining implementations viz. GS--3, GS--5 and GS--6 have yielded slightly improved F-Measures than that of Sauvola. This shows that the results with the proposed implementations are comparable with the result of Sauvola. Fig. \ref{fig_benchmark_bnz} show some sample images of the above datasets and their binarized images for some techniques.

%
%
%

\subsection{Computational Complexity and Computation Time}
\begin{figure}
  \centering
  \includegraphics{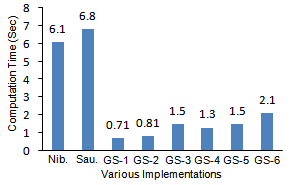}
  \caption{Plot of mean computation times of Niblack, Sauvola and proposed techniques (for the images of resolution 1024x768 with 20x20 window size)}
  \label{img_comp_time}
\end{figure}

The proposed technique calculates the threshold for each pixel with computation time of $O(W)$ time. So, computational complexity of the proposed technique is $O(WN^2)$. Plot of mean computation times of Niblack, Sauvola and proposed techniques has been shown in Fig. \ref{img_comp_time} with respect to a moderately powerful notebook (DualCore T2370, 1.73 GHz, 1GB RAM, 1MB L2 Cache). It may be observed from Fig. \ref{img_comp_time} that the computation time of the proposed technique is much lesser than Niblack's and Sauvola's implementations.

\subsection{Memory Consumption}
As storing a pixel of a gray scale image requires 1 byte of memory, an $N\times N$ image requires $S_z=N\times N$ bytes of memory. The algorithmic implementation of Niblack's and Sauvola's technique makes a copy of the image before binarizing its pixels by convolving the window. So, the amount of memory consumption of this algorithm is $2\times S_z+c_1$ bytes where $c_1$ is a constant.

The implementation proposed by Shafait et al. \cite{Shafait_Eff} is faster, but it requires additional  memory. It prepares two types of integral images from the given image - one for intensity values and the other for square of the intensity values. To store these integral images with 32 bit and 64 bit integers respectively, we need $4\times S_z$ and $8\times S_z$ bytes of memory. So, the amount of memory consumption in this case is $12\times S_z+c_2$ where $c_2$ is another constant. It may be noted that the memory consumption is 6 times higher than that of the algorithmic implementation.

 Memory consumption of our implementation can be given as $2\times S_z+c_3$ where $c_3$ is another constant. It may be noted that this implementation requires no additional memory compared to the original/algorithmic implementation.

\section{Conclusion}
In this paper, we have presented a novel implementation of convolution based locally adaptive binarization techniques. Both the computational complexity and computation time are significantly reduced while keeping the performance close to the ordinary implementation. The computational complexity has been reduced from $O(W^2 N^2)$ to $O(WN^2)$ and the time computation has been reduced by 5 to 15 times depending on the window size. At the same time, memory consumption is the same with the original implementation. This type of implementation is especially useful in image analysis and document processing systems for real-time systems and on handheld mobile devices having limited computational facilities. As the trend in designing camera based applications on mobile devices has recently increased considerably, the presented technique will be highly useful.

\subsubsection{Acknowledgments.}
We are thankful to the \emph{Center for Microprocessor Application for Training Education and Research (CMATER)}, Jadavpur University for providing infrastructural support for the research work. The first author is also thankful to the \emph{School of Mobile Computing and Communication (SMCC)} for providing fellowship to him.

\end{document}